# BOUNDARY REGULARIZED BUILDING FOOTPRINT EXTRACTION FROM SATELLITE IMAGES USING DEEP NEURAL NETWORKS

Kang Zhao, Muhammad Kamran, Gunho Sohn

Department of Earth and Space Science and Engineering, Lassonde School of Engineering
York University, Canada. kangzhaogeo@gmail.com, (mkamran9, gsohn) @yorku.ca

**ABSTRACT:**

In recent years, an ever-increasing number of remote satellites are orbiting the Earth which streams vast amount of visual data to support a wide range of civil, public and military applications. One of the key information obtained from satellite imagery is to produce and update spatial maps of built environment due to its wide coverage with high resolution data. However, reconstructing spatial maps from satellite imagery is not a trivial vision task as it requires reconstructing a scene or object with high-level representation such as primitives. For the last decade, significant advancement in object detection and representation using visual data has been achieved, but the primitive-based object representation still remains as a challenging vision task. Thus, a high-quality spatial map is mainly produced through complex labour-intensive processes. In this paper, we propose a novel deep neural network, which enables to jointly detect building instance and regularize noisy building boundary shapes from a single satellite imagery. The proposed deep learning method consists of a two-stage object detection network to produce region of interest (RoI) features and a building boundary extraction network using graph models to learn geometric information of the polygon shapes. Extensive experiments show that our model can accomplish multi-tasks of object localization, recognition, semantic labelling and geometric shape extraction simultaneously. In terms of building extraction accuracy, computation efficiency and boundary regularization performance, our model outperforms the state-of-the-art baseline models.

## 1. INTRODUCTION

Acquiring information about the structure on the surface of the earth without making physical contact is generally achieved by the remote sensing techniques (Campbell, Wynne, 2011). Applications like digital mapping, land use analysis, disaster monitoring and climate modelling largely use satellite images. The satellite images have been important in the creation of the digital maps for the Geographic Information System (GIS) and the building footprint information is playing an instrumental role in urban planning, smart city construction and many others. In addition, the building footprints with regularized boundaries are able to produce polygons of vector representation, which hold stronger transferability over multiple GIS platforms therefore having an expensive domain of applications. For example, the regularized building polygons can produce more accurate 3D building models. Nonetheless, as satellite images are readily available and accessible, so there is always a demand for better quality of the building footprints. This demand has not yet been properly fulfilled due to numerous challenges. Firstly, the building footprints on the GIS maps need the manual or semi-automatic procedure to reach the high precision, which is quite time-consuming and labour-intensive. Secondly, the enormous diversity of the outlooks of the building roofs creates barriers for large-scale building footprint extraction. Also, the geometric potential of the satellite images has not been exploited. Due to the pixel-wised and grid-based representation of the images, it's fairly demanding to learn the geometric information of polygon shapes. In recent years, deep learning has brought a revolution in Artificial Intelligence (AI).

Deep learning is largely been used in the fields of computer vision, speech recognition, natural language processing, etc and it can give some exceptional results as compared to many traditional techniques in the remote sensing domain as well.

Motivated by the challenges of regularized building footprint extraction and to exploit the potential of deep learning, in this study, we will present our model which is based on deep neural networks. The model utilizes the spatial, semantic and geometric information to perform automatic building footprint extraction and handle the problem of building boundary regularization.

Our general framework illustrated in Fig 1, takes advantage of the typical supervised learning mechanism and can be considered as the instance segmentation model combined with geometric learning. Our framework can simultaneously recognize and localize multiple objects, assign semantic labels at pixel-level and predict polygons of geometric shapes.

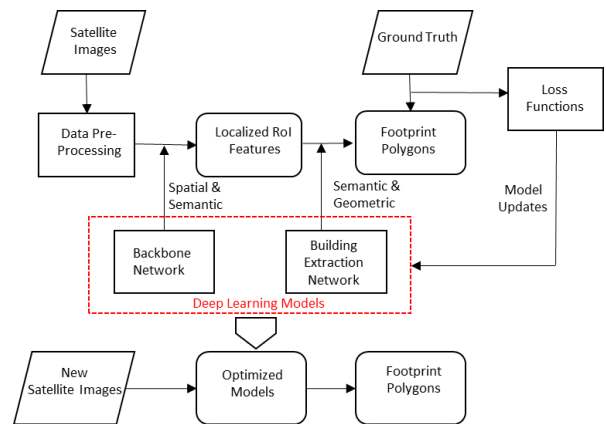

Figure 1. Framework of the Study

## 2. RELATED WORK

Deep learning is a special type of machine learning algorithm that follows a multi-layer structure to learn data representation or features with multiple levels of abstraction (LeCun et al., 2015). The traditional machine learning techniques tend to use hand-crafted features for the data representation. However, deep neural networks employ a combination of both linear and non-linear operations to encode deep features from input data. In the supervised learning scenario, the network outcomes are compared with the ground truth through the loss function. The deep learning models for object detection can be classified into two-stage and one-stage models.

Two-stage models generally utilize a dual-stage detection pipeline. The Region-based Convolutional Neural Network (R-CNN) is the meta-model for the two-stage models. The R-CNN model (Girshick et al., 2014), adopts a search algorithm to produce about 2000 region proposals from an image and feed them into CNN to extract features. Then a support vector machine (Cortes, Vapnik, 1995) is used to classify the regions and predict bounding boxes based on the extracted features. Fast R-CNN (Girshick, 2015) initially inputs the image into CNN to extract a feature map and crop the region proposals with the feature map to generate the region of interest (RoI) features. For localization and classification, it employs the fully connected layer and the softmax function (Bridle, 1990). Compared to R-CNN, Fast R-CNN applies CNN over the whole image once. However, both of them need a pre-processing step to generate the region proposals and takes too much time. To address this problem, Faster R-CNN (Ren et al., 2015) abandons the search algorithm and designs a Region Proposal Network (RPN) to generate proposals from pre-defined anchor boxes.

One-stage models skip the stage of region proposal generation and directly apply the one-shot detection over densely sampling possible locations of the input image. One-stage models has an edge over two stage models due to their simplified and unified network design. For instance, YOLO (Redmon et al., 2016) splits an image into fixed size of grids, on each of which the CNN is applied to predict the bounding boxes and class probabilities. SSD (Liu et al., 2016) additionally employs a series of convolution layers with decreasing sizes to extract the pyramid of multi-scale features, on which objects with different sizes can be detected. Recent works like (Law, Deng, 2018) and (Zhou et al., 2019) accomplish the object detection by directly using CNN to detect representative key points of objects, from which the bounding box predictions can be produced.

Image segmentation is the core issue in our study and huge amount of efforts have been invested in this area by deep learning researchers. There are two types of segmentation research domains: semantic segmentation and instance segmentation.

Two approaches are generally adopted to achieve instance segmentation. One is to first perform semantic segmentation over the image and then apply instantiation by grouping the connected pixels to identify individual objects. This pipeline is utilized by DeepMask (Pinheiro et al., 2015) and SharpMask (Pinheiro et al., 2016). The other approach is put forward by Mask R-CNN (He et al., 2017). It first performs instantiation and then segmentation. The object detection network (almost same with Faster R-CNN) is employed to distinguish and localize objects. The detection part can also generate well-localized Region of Interest (RoI) features, over which the semantic segmentation model such as FCN is applied to obtain object masks. The whole network has an end-to-end unified design. The segmentation accuracy of Mask R-CNN surpasses the models adopting the first approach on most of the benchmarks. Early works (Mnih, 2013) trained a basic CNN for building labelling, which only contains three layers including one convolution layer, one pooling layer and one fully connected layer. It shows competitive results compared to other complicated traditional algorithms, but the simple CNN is quite sensitive to the hyperparameter setting. More recent works employ more complex CNN models. (Maggiori et al., 2017) designed a multi-layer perceptron structure, which has a skip connection similar to the U-Net to combine features of different scales. The SegNet is directly used by (Bischke et al., 2017) to train an additional loss representing the distance to the building boundary apart from the pixel-wised classification loss. (Wu et al., 2018a) utilizes the U-Net as the basic model with multiple constraints, which restrict the outputs from feature maps of different scales to be compared with the ground truth images of corresponding scales. Other works adopt the data-fusion idea to boost the segmentation performance and still use the semantic segmentation models to deal with data of multi-sources. LiDAR point clouds and images are combined in (Liu et al., 2017) through a U-Net model. (Audebert et al., 2017) employ the U-Net architecture to utilize the satellite images and GIS maps like OpenStreetMap and Google Map to take advantage of the more precise vectorized maps. (Pan et al., 2019) employs the U-Net enhanced by the GAN model with spatial and channel attention mechanisms to produce more discriminative prediction maps.

The application of the Mask R-CNN is explored in (Zhao et al., 2018) for the building extraction problem and achieves a satisfying instance segmentation performance. (Wen et al., 2019) further improves the Mask R-CNN model by introducing the rotational bounding boxes to enhance detection quality and stacking the receptive field blocks to handle scale viability issues. The building boundary regularization is one of the centre problems of our study, which is typically associated with geometric learning of polygon shapes. Before the deep learning era, the building footprint extraction relied more on the processing of the LiDAR point clouds than the images because the point clouds hold the spatial locations of the points, which are more geometrically meaningful. (Jung et al., 2017) and (Jung et al., 2019) mainly use point cloud data and adopt a Binary Space Partitioning (BSP) process and a Minimum Description Length (MDL) based algorithm to generate and optimize the building polygon shapes for building footprint detection and boundary regularization. To fully exploit the boundary information, (Marmanis et al., 2018) feeds the fusion of the images and Digital Elevation Model (DEM) to the SegNet model combined with extra edge and boundary predictions produced from the FCN model and adopt a multi-task learning strategy. (Volpi, Tuia, 2018) and (Wu et al., 2018b) also utilize the multi-task learning scheme to train additional boundary loss on FCN or U-Net models, and they claim that they can produce building boundaries with regularities. Moreover, some conventional polygonal models such as the active contour (ACM) or snake model (Kass et al., 1988) are recycled in the modern CNN architectures. The deep structured active contours (DSAC) (Marcos et al., 2018) and the deep active ray network (DARNet) (Cheng et al., 2019) both integrate the ACM model into their segmentation networks to learn richer geometric information to better predict the polygon contours and delineate the building boundaries

There are also deep learning models trying to directly generate polygons instead of pixel-wised segmentation maps. Many of them do so by producing the optimal locations of the polygon vertices and linking the predicted vertices with straight lines, which will intuitively produce polygons with regularized boundaries. PolyRNN (Castrejon et al., 2017) and PolyRNN++

(Acuna et al., 2018) employ the recurrent neural networks to predict the locations of polygon vertices in sequence, that is, the current vertex prediction is influenced by the previous predictions. These two models are applied for semi-automatic annotation with bounding boxes provided, thus failing to produce object detection results in their frameworks. (Li et al., 2018) borrows the ideas of PolyRNN and Mask R-CNN to build a unified pipeline to accomplish object detection and sequential polygon vertex prediction and applies the framework on large-scale image datasets to extract building footprints and road lines. CurveGCN (Ling et al., 2019) explores the usage of the graph convolutional networks (GCN) to produce polygons as a graph representation, which is efficient and utilize more geometric features than RNN models. However, like PolyRNN and PolyRNN++, CurveGCN is also used for annotation tasks and is unable to perform object detection.

## 3. BACKBONE NETWORK

The backbone network is designed for feature encoding and building object detection and localization. We utilize a combination of Residual Network (He et al., 2016) and Feature Pyramid Network (Lin et al., 2017) to extract deep features at multiple scales. We believe that detecting objects at different scales is essential since our input satellite images cover lager area of lands and contain many building objects with various sizes. The multi-scale feature maps obtained from the feature encoding network are capable of recognizing building objects from different scales compared to those using feature maps of only one scale. To detect and localize building objects, a two-stage object detection model is employed including the Region Proposal Network (RPN) and a localization layer, including bounding box regression and classification layers. The RPN first takes the features and pre-defined anchor boxes to generate the initial proposed bounding boxes, which are used to crop with the feature maps to get the cropped features. In the cropping process, we pick the feature maps to crop based on the size of the box proposals following the Eq 1.

$$k = [k_0 + log_2\sqrt{wh}/224]  \qquad (1)$$

where w, h are the width and height of the box proposal; $k_0= 4$ and k is the level of scale we select as the feature map to crop. Since the cropped features have various sizes, we feed them into a RoI pooling layer, which is then operated on the cropped features to obtain RoI features. These features are fed into the box regression and classification layer to produce the coordinates and class scores of the refined bounding boxes. Lastly, the multi-scale feature maps and the final bounding boxes are input into the RoI-Align layer to generate precisely localized RoI features. The localized RoI features play a critical role in other tasks like pixel-wised segmentation or geometric shape learning. The whole structure of the backbone network is illustrated as Fig 2.

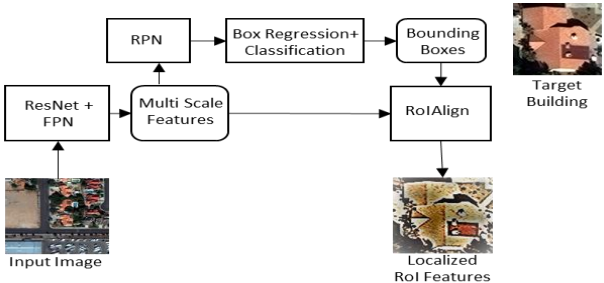

Figure 2. Backbone Network

### 3.1 Loss Design for Backbone Network

For the backbone network, we need to calculate losses for the object detection of two stage. The loss functions for both stages deal with two types of losses, box regression loss and classification score loss, thus forming a multi-task training scenario. Following Faster R-CNN, the box deltas are calculated as inputs of the box regression loss function rather than box coordinates. The deltas are defined as

$$t_x = (x - x_a)/w_a, t_y = (y - y_a)/h_a,$$
$$t_w = log(w/w_a),  t_h = log(h/h_a) \qquad (2)$$

$$t_x^* = (x^* - x_a)/w_a, t_y^* = (y^* - y_a)/h_a,$$
$$t_w^* = log(w^*/w_a),  t_h^* = log(h^*/h_a) \qquad (3)$$

In the Eq 2, $x, y$ represents the coordinates of the centre point of the bounding box while $w$ and $h$ represent its width and height. Respectively, $x$, $x_a$ and $x^*$ are for the predicted box, anchor box and ground truth box (likewise for $y$, $w$ and $h$). Our backbone network will predict the box deltas ($t_x, t_y, t_w, t_h$), which are equivalent to the regression from a pre-defined anchor box to a predicted box. To compare the predicted box deltas with the ground truth ones ($t_x^*, t_y^*, t_w^*, t_h^*$), which represent the regression values from an anchor box to its closest ground truth box, we adopt the smooth L1 loss:

$$L_{reg}(t, t^*) = \sum_{i \in x,y,\omega,h} smooth_{L1}(t_i - t_i^*) \qquad (4)$$

Where a $smooth_{L1}$ loss is:

$$smooth_{L1}(x) = \begin{cases} 0.5x^2 & if|x| < 1 \\ |x| - 0.5 & otherwise \end{cases} \qquad (5)$$

As it's a binary classification problem, we compute a binary cross entropy loss:

$$L_{cls}(p(y)) = -(y \, log(p(y)) + (1-y) \, log(1 \, p(y))) \qquad (6)$$

Here, $y$ is the class label predicted (0 or 1) and $p(y)$ is the probability score for the class label.

When training RPN, the anchor boxes are assigned class labels as positive or negative, which we refer as objectness scores. The box proposals are also generated as a coarse localization result. Therefore, based on Eq 4 and Eq 6, we define RPN loss as:

$$L_{rpn}(p^{obj}, t^{rpn}) = \frac{1}{N_{cls}} \Sigma_{i=1}^{N_{cls}} L_{cls}(p_i^{obj}(y)) +$$
$$\frac{1}{N_{box}} \Sigma_{i=1}^{N_{box}} L_{reg}(t_i^{rpn}, t_i^*) \qquad (7)$$

where $p^{obj}$ and $t^{rpn}$ are the objectness scores and proposed box deltas by RPN while $N_{cls}$ and $N_{box}$ are the number of all the boxes (both positive and negative) after NMS and the number of the positive boxes. As negative boxes contain no object, the box deltas of positive ones are used for box regression. Also, we need to balance the classification loss and box regression loss with weight coefficients. To address this issue, we simply add the two losses without using any coefficients, which is similar to Faster R-CNN.

The localization layer classifies the boxes into two classes, building and non-building and produces final location predictions of the bounding boxes. For these two tasks, we compute a localization loss and a similar loss function design is adopted:

$$L_{loc}(p^{class}, t^{loc}) = \frac{1}{N_{cls}} \Sigma_{i=1}^{N_{cls}} L_{cls}(p_i^{class}(y)) + \frac{1}{N_{box}} \Sigma_{i=1}^{N_{box}} L_{reg}(t_i^{loc}, t_i^*) \quad (8)$$

where $p^{class}$ and $t^{loc}$ are the class scores and the predicted bounding box deltas.

## 4. REGION BASED POLYGON GCN (R-POLYGCN)

Even integrated with polygon shape-prior, the previous models are no more than traditional segmentation methods, which are intended to label every pixel of the images and are not capable of directly exploiting the geometric shape information. It is due to the fact that the pixel-wised representation of the polygon shape has much less geometric meaning than using vertices and edges to delineate polygons. However, graph model is exactly a representation of data structure with vertices and edges, which can be employed to depict our building polygons with much richer geometric property. The graph neural network (GNN) can also be designed for convolution operations to allow the feature information exchange among vertices for geometric learning. Hence, to leverage the geometric nature of the graph model and the GNN, we combine the object detection backbone network and a graph convolutional network to propose our R-PolyGCN, a region based GCN to specially detect buildings from satellite images and directly predict the locations of polygon vertices by implicitly learning the geometric polygon shapes. Then simply joining the vertices in a sequence with straight lines will result in building footprint extractions with more regularized boundaries.

### 4.1 Network Architecture

The architecture of R-PolyGCN is illustrated in Fig 4.1. Initially, the backbone network is utilized to detect the target buildings and provide well-localized RoI features. Secondly, the geometric shapes of building polygons are learned through GCN on these well-localized regions. From the RoI features, we additionally predict boundary masks and concatenate them as boundary features onto the RoI feature map to obtain enhanced features.

Since the major goal of our GCN is to move the initial polygon vertices to the boundary of the target building polygons, we first generate initial vertices, which follow a pre-defined order. Then initial graph is generated and the graph features for each vertex are interpolated from the enhanced features. Afterwards, graph features are fed into a multi-step and multi-layer graph convolutional networks, which can predict the vertex offsets. By adding the vertex offsets to the initial vertex coordinates and connecting the vertices we can acquire the final polygon prediction.

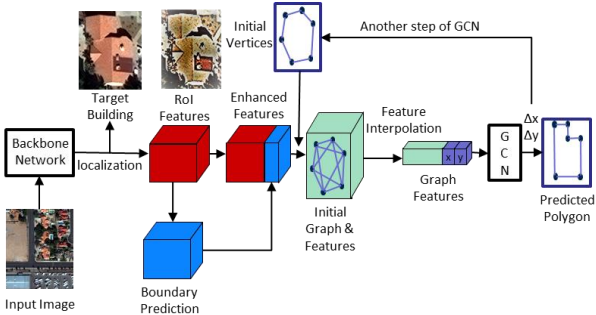

Figure 4.1: Network Diagram of R-PolyGCN

On top of the RoI feature maps extracted from the backbone network, we specially trained two fully connected layers to predict polygon boundary masks including edge masks and vertex mask of the target building. The boundary prediction represents the pixel-wised probabilities of edges and vertices of the building polygon. The edge logits and vertex logits of the predicted boundary are then concatenated with RoI features to create an enhanced feature map $F_{en}$. We believe that the enhanced features can outperform plain RoI features in terms of recognizing building boundaries because of their confidence of polygon boundary existence.

The polygon vertices of the target building are initialized using N points, which are allocated as the vertices of a regular polygon. Linking the initial vertices with straight lines generates the initial polygon. The number of vertices per polygon is unified and kept the same with the ground truth data. Also, to make the sequence of vertices well defined and the topology of the polygon well reserved, these vertices are kept in clockwise order. Then the initial vertices are put at the central part of the enhanced feature map $F_{en}$. Let $V_i = [x_i, y_i]$ denote the location of the i$^{th}$ vertex and $V = (v_1, v_2, \ldots, v_N)$ be the set of all polygon vertices, which serve as the initial nodes of our graph model. The edges of the graph E are produced by connecting each node in V with its four neighbouring nodes. Linking nodes of the graph in this way allows five neighbouring nodes to exchange information and affect each other in GCN, which means longer geometric coherence. Lastly, we define the initial graph as $G = (V, E)$. For one node $v_i$, based on its location in $F_{en}$, the bilinear interpolation is adopted to obtain its node features $F_{en}(x_i, y_i)$ from the enhanced features. Then we concatenate the node's current location $(x_i, y_i)$ and its node features in the following way:

$$f_i = concat\{F_{en}(x_i, y_i), x_i, y_i\} \quad (9)$$

where $f_i$ is the graph feature for the node $v_i$ and will be input into GCN. So, the input graph features for each vertex are a combination of the enhanced features and the vertex location information.

We employ a multi-step architecture here to achieve a coarse-to-fine polygon prediction. At the first step, the initial graph features are fed into a GCN to get first-round initial offsets of the polygon vertices. Then we adjust the locations of the vertices by the predicted offsets and obtain new graph features interpolated from the enhanced features again. Then the adjusted vertices are fed into another GCN and produce another vertex offset in the second step. The procedure will be iterated in the following steps so that we can get more and more accurate vertex locations as well as polygon prediction. In this work, we adopt a three-step GCN. Within each step, a multi-layer GCN is adopted. Assuming that $N(v_i)$ denotes the parameters at the layer l, the basic graph convolution calculation for the node $v_i$ at this layer is defined as:

$$f_i^{l+1} = w_0^l f_i^l + \Sigma_{v_j \in N(v_i)} w_1^l f_j^l \quad (10)$$

in which $f_i^l$ is the graph feature for the node $v_i$ and $f_i^{l+1}$ is the output of the convolution operation at layer l. Instead of the basic convolution operations, we utilize a residual block from ResNet, which is displayed in Fig 4.2.

The computations of the residual block are formulated as:

$$r_i^l = RELU(w_0^l f_i^l + \Sigma_{v_j \in N(v_i)} w_1^l f_j^l) \quad (11)$$

$$r_i^{l+1} = (\tilde{w}_0^l r_i^l + \Sigma_{v_j \in N(v_i)} \tilde{w}_1^l r_j^l) \quad (12)$$

$$f_i^{l+1} = RELU(r_i^{l+1} + f_i^l) \quad (13)$$

Both Eq 11 and Eq 12 are two-layer graph convolutions similar to Eq 10 but are aimed to output the residual $r_i^{l+1}$. The convolution weights are $w_0^l, w_1^l, w_0^{\tilde{l}}$ and $w_1^{\tilde{l}}$. Then we add the residual $r_i^{l+1}$ and the identity of the input $f_i^l$. After a ReLU activation layer, the final output $f_i^{l+1}$ is obtained.

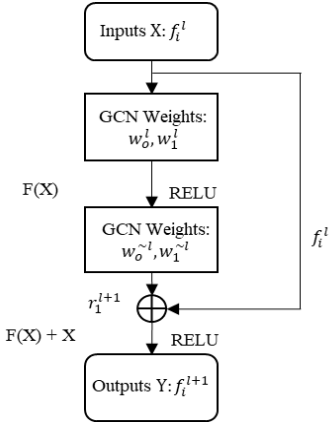

Figure 4.2: GCN: Residual Block

**4.2 Loss Design for R-PolyGCN**

To accomplish objection detection for well-localized regions of interest, we still need to train the RPN loss and localization loss described in section 3.1. The rest of losses are for the boundary prediction and polygon vertex prediction.

**4.2.1 Boundary Prediction Loss**: Both vertex masks and edge masks are binary and are produced to enhance the features. So, the binary cross entropy loss function is applied to compute the vertex mask loss $L_{v-mask}$ and edge mask loss $L_{e-mask}$:

$$L_{v-mask}(p^v(y)) = -\frac{1}{N}\Sigma_{i=1}^N (y \log(p_i^v(y)) + (1-y) \log(1-p_i^v(y))) \quad (14)$$

$$L_{e-mask}(p^e(y)) = -\frac{1}{N}\Sigma_{i=1}^N (y \log(p_i^e(y)) + (1-y) \log(1-p_i^e(y))) \quad (15)$$

where $p^v(y)$ and $p^e(y)$ are the pixel-wised probability of vertex mask and edge mask; y is the binary class label 0 or 1 while $N$ is the total number of pixels. Therefore, we can have the boundary prediction loss $L_{boun}$:

$$L_{boun} = L_{e-mask} + L_{v-mask} \quad (16)$$

**Polygon Localization Loss:** A polygon vertex location is denoted by its coordinates $v(x, y)$ and a polygon location is represented by its N vertices: $p = \{v_i \mid i = 1,2, \ldots, N\}$ Assume our model has extracted K polygons, which are $P = \{p_k \mid k = 1,2,\ldots,K\}$ The polygon vertices are defined in clockwise order like the ground-truth vertices. Both point sets have the same number of vertices per polygon as well. $p^{pre}$ by using the geometric L$_1$ distance, which is defined as:

$$L_1(p^{pre}, p^{gt}) = \Sigma_{i=0}^N (|x_i^{pre} - x_i^{gt}| + |y_i^{pre} - y_i^{gt}|) \quad (17)$$

However, the vertex correspondences aren't matched between these two-point sets since the starting vertices are unknown. To find such correspondences, we fix the starting vertex of ground truth point sets and adopt an exhaust search for the optimal corresponding starting vertex of the predicted sets, which means the predicted point sets will be expanded by using every vertex as the starting one. For one polygon, assume that the number of the vertices per polygon is N. Then N different predicted point sets are generated from original point set for the polygon. These point sets have the same clockwise order but N different starting vertices. The L$_1$ distances will be calculated between the ground truth point sets and all of N expanded predicted point sets, thus resulting in N polygon distances. Among these distances, the smallest one will be selected as the polygon localization loss and the optimal correspondence of vertices can also be found. Taking all K extracted polygons into account, we have K ground truth polygons and the number of predicted polygons will be expanded to K × N. Then the loss function for all the K polygons can be formulated as:

$$L_{poly}(p^{pre}, p^{gt}) \frac{1}{K}\Sigma_{k=1}^K min_{j \in (0,1,\ldots,N-1)}(L_1(p_{k+j}^{pre}, p_k^{gt})) \quad (18)$$

where $L_{poly}$ denotes our polygon localization loss, which is an average polygon distance with vertex correspondences. Overall, the total loss function for our R-PolyGCN is:

$$L_{R-PolyGCN} = L_{rpn} + L_{loc} + L_{boun} + L_{poly} \quad (19)$$

**4.3 Training Strategy**

Training all the losses of R-PolyGCN is challenging and we provide some training strategies. Most of our losses are either entropy loss or smooth L$_1$ loss, which can be well trained in parallel. However, the reality of the polygon localization loss is geometric point distance, which leads to obstacles when training with other losses for several reasons. Firstly, the geometric point distance has various scales and is not normalized. On the other hand, feasible and stable polygon localization loss can be only obtained until the target building regions are stabilized, which happens when the RPN loss and localization loss are small and stable. Before that, due to the region localization is not fully optimized in the backbone network, incomplete building polygons with less geometric meaning will be generated. Because our GCN model relies on the polygon geometric features, the polygon localization loss will become unreasonable causing the GCN not well trained or even wrongly trained. To tackle the training obstacles, the following strategies are utilized.

**Polygon localization loss normalization**: We first restrict the coordinates of the polygon vertices to [0, 1]. The polygon distance is then divided by the vertex number of the polygon. A weight coefficient λ is added to the polygon localization loss as well. Then our new loss function is:

$$L_{R-PolyGCN} = L_{rpn} + L_{loc} + L_{boun} + \lambda \frac{1}{N} L_{poly} \quad (20)$$

where N is the number of vertices per polygon. λ will be considered as a hyper parameter and referred as polygon localization weight. The parameter is set to belong to {0.5, 0.75, 1.0, 1.25} and will be fine-tuned during training. In this way, the polygon localization loss is regularized to same scale as other losses, which allows balanced losses to be trained.

The training is divided majorly into three stages. At the early stage of the training, we only train the RPN & localization layer in the backbone network and the boundary prediction part and "freeze" the GCN polygon vertex prediction part. It is intended to obtain stably localized regions for the target buildings by only training the backbone network. After certain epochs of training, the GCN part begins to be trained to optimize the polygon prediction while keeping the backbone network frozen. Finally, the GCN part and the backbone network are trained together to fine tune the whole network. By adopting the multi-stage training pipeline, the negative effects that the backbone network can possibly bring to the GCN polygon prediction can be avoided.

## 5. EXPERIMENTS AND RESULTS

We utilize the open dataset provided by the building extraction challenge of DeepGlobe workshop (Demir et al., 2018) at CVPR, 2018. The data contain high-resolution satellite images and ground truth for the building footprints. The study area of this dataset consists of four cities (LasVegas, Paris, Shanghai, Khartoum) and covers both urban and suburban regions. The images are captured by the DigitalGlobe Worldview-3 Satellite with GeoTiff data format. The image size is 650×650 and the resolution is 30cm, which allows the image to cover regions of 200m × 200m area. The workshop allowed 10,593 images with labelled files for public use. The dataset information is displayed in the table 5.1

| City | Area (km$^2$) | Building Annotation | Number of Images | Data Amount (GB) |
|---|---|---|---|---|
| Vegas | 216 | 108,328 | 3851 | 23 |
| Paris | 1030 | 16,207 | 1148 | 5.3 |
| Shanghai | 1000 | 67,906 | 4582 | 23.4 |
| Khartoum | 765 | 25,046 | 1012 | 4.7 |
| Total | 3011 | 217,487 | 10,593 | 56.4 |

Table 5.1: Information of the dataset

The experimental data was acquired from Amazon Web Service with licence from the DeepGlobe workshop. The dataset provides several types of satellite images, from which we selected the Pan-sharpened RGB images for our experiments. Before inputting the images into our deep learning models, the data was pre-processed. Our neural network models, training and inference codes were implemented with Python 3.6 on Pytorch 0.4.0, an open source deep learning platform. The configurations of the backbone network are displayed in the table 5.2

| | Items | Configuration |
|---|---|---|
| Feature Encoding | Input Image Size | (1024,1024) |
| | ResNet Layers | Res-101 |
| | FPN Feature Sizes | (32, 32), (64, 64), (128, 128), (256, 256) |
| RPN | Anchor Stride | [4,8,16, 32,64] |
| | Anchor Shape | [0.5, 1, 2] |
| | Anchor Scale | [32, 64, 128, 256, 512] |
| | NMS Threshold | 0.5 |
| | Max Box Number | 256 |
| | Positive/Negative Ratio | 1:3 |
| Localization layer | RoI Size | (28, 28) |

Table 5.2: Configuration of Backbone Network

The key parameters for our R-PolyGCN model are summarized in the table 5.3, in which 1+4 means that one node is connected with four neighbouring nodes in the graph.

| Items | Configurations |
|---|---|
| Number of Vertices per Polygon | 16 |
| Vertex Order | Clockwise |
| Connected Nodes in the Graph | 1+4 |
| GCN Steps | 3 |

Table 5.3: Configuration of R-PolyGCN Model

The results are qualitatively analysed to demonstrate the properties of our models and to prove that they are able to conquer the challenges of the building extraction mentioned in the section 1 including Automatic extraction procedure, Handling the diversity of building roof outlooks, Balance between recognition and localization, Distinguishing closely located buildings, Detecting buildings of various sizes and Capturing geometric shapes of polygons.

**Building Extraction Accuracy**: We took advantage of the accuracy metrics provided by the DeepGlobe workshop, which were computed by comparing the locations of the predicted building polygons and the ground truth ones. Firstly, we utilized the metric of Intersection over Union (IoU). The predicted building polygon was counted as a true positive if it was the closest (measured by the IoU) proposal to a labelled polygon and the IoU between the prediction and the label was beyond the prescribed threshold of 0.5. Otherwise, the proposed polygon was a false positive. The labelled polygons that were not detected or missed in the predictions were denoted as false negative. After counting true positives (TP), false positives (FP) and the false negative polygons (FN), we employed the F1-score. As compared to the state-of-the-art instance segmentation model, Mask R-CNN, our model, R-PolyGCN consistently had the highest detection accuracy over all other models and on the dataset of all cities. Note that the relatively low F1-scores for Shanghai and Khartoum result from the annotation of lower quality. Because our dataset was acquired from the open challenge by the SpaceNet building dataset and DeepGlobe workshop, many participants had produced their results which were recorded and ranked on the public leader boards. For evaluation, we can only submit the outcomes onto their online evaluation system to obtain the results. We compared the evaluation results of the F1-scores of our models with other top participants in the table 5.4.

| Models or Participants | F1 Score (Individual City) | | | | Total F1 Score |
|---|---|---|---|---|---|
| | Las Vegas | Paris | Shan-ghai | Khart-oum | |
| Wleite | 0.829 | 0.679 | 0.581 | 0.483 | 0.643 |
| XD_XD | 0.885 | 0.745 | 0.597 | 0.544 | 0.693 |
| Mask R-CNN | 0.881 | 0.760 | 0.646 | 0.578 | 0.717 |
| Mask R-CNN (Regularized) | 0.879 | 0.753 | 0.642 | 0.568 | 0.713 |
| R-PolyGCN | **0.892** | **0.786** | **0.682** | **0.612** | **0.742** |

Table 5.4: F1-scores of Building Extraction Results

Other participants include the top 2 winners of the SpaceNet competition, Wleite and XD_XD. Both of these participants adopted the semantic segmentation models (SS) followed by post-processing algorithms (PP) and our approach was the instance segmentation models (IS) with geometric learning for polygon shapes. From the table 5.4, it is proved that our R-PolyGCN outperformed all of them.

**Building Regularization Performance:** The performance of the building boundary regularization was evaluated. Since Mask R-CNN outputted pixel-wised segmentation results without vertex or line prediction, we used a contour tracing algorithm in OpenCV to obtain the boundaries from the masks. In terms of the regularization of building boundaries, we can observe that Mask R-CNN shows almost no evidence to regularize the boundaries because of its nature of grid-based pixel-by-pixel representation and lack of shape information. But R-PolyGCN models can produce boundary lines closest to the ground truth with regularized characteristics as shown in Fig 5. In the figure, examples of building footprint extraction with focus on the boundaries are shown where red points represent vertices and green lines represent polylines.

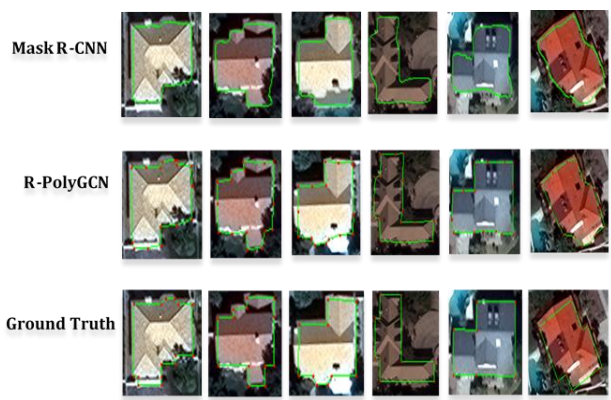

Figure 5: Comparison of the performance of building boundary regularization.

For R-PolyGCN, as a natural representation for the vertex, edge and polygon, the graph model employed can provide a straightforward polygon prediction based on their geometric features. Once the optimal polygon vertices are acquired, connecting them by straight lines in a pre-defined order can easily produce regularized boundaries.

## CONCLUSION

In this study, we aimed to develop a deep learning framework to automatically extract building footprints with boundary regularization from satellite images. We systemized the main problems into the tasks of spatial learning, semantic learning and geometric learning and proposed a general deep learning-based framework. To provide the solution for the boundary regularized building footprint extraction, our methodology is composed of the backbone network and the building extraction network. The backbone network has the functions of multi-scale feature encoding and object detection while also producing the well-localized RoI features. We explored R-PolyGCN as building extraction network, which exploits the graph representation for polygons and the graph convolutional networks for geometric learning. Particularly, our R-PolyGCN outperformed all the others in terms of extraction accuracy. The efficiency of the model is also analysed, which showed that R-PolyGCN is the most efficient one at both training and inference stages. We compared the performances of model on the building boundary regularization and find that our R-PolyGCN demonstrate outstanding capacity to produce regular building boundaries.

## ACKNOWLEDGEMENT

We would like to acknowledge the supports from Natural Science and Engineering Research Council of Canada (NSERC) Discovery Program and Intelligent Systems for Sustainable Urban Mobility (ISSUM) Program.